\documentclass[10pt]{article} 
\usepackage[preprint]{tmlr}

\usepackage{hyperref}
\usepackage{url}

\usepackage{amssymb}
\usepackage{graphicx}
\usepackage{amsmath}
\usepackage{booktabs}       
\newcommand{\modelname}{\textsc{EHRDiff} }
\usepackage{algorithm}
\usepackage{algorithmic}

\usepackage{mathtools}
\usepackage{microtype}
\usepackage{inconsolata}

\usepackage{tmlr}

\usepackage{amsmath,amsfonts,bm}

\def\eqref#1{equation~\ref{#1}}

\def\1{\bm{1}}

\DeclareMathAlphabet{\mathsfit}{\encodingdefault}{\sfdefault}{m}{sl}
\SetMathAlphabet{\mathsfit}{bold}{\encodingdefault}{\sfdefault}{bx}{n}

\newcommand{\E}{\mathbb{E}}

\title{\modelname: Exploring Realistic EHR Synthesis with Diffusion Models}

\author{\name Hongyi Yuan$^\dag$ \email yuanhy20@mails.tsinghua.edu.cn \\
      \addr Center for Statistical Science \\
      Tsinghua University
      \AUAND
      \name Songchi Zhou$^\dag$ \email zhou-sc23@mails.tsinghua.edu.cn \\
      \addr Center for Statistical Science \\
      Tsinghua University
      \AUAND
      \name Sheng Yu\thanks{Corresponding Author. $\dag$ Contributed Equally. Codes are released in \url{https://github.com/sczzz3/EHRDiff.git}.} \email syu@tsinghua.edu.cn\\
      \addr Center for Statistical Science \\
      Tsinghua University}

\def\month{03}  
\def\year{2024} 
\def\openreview{\url{https://openreview.net/forum?id=DIGkJhGeqi}} 

\begin{document}

\maketitle

\begin{abstract}
Electronic health records (EHR) contain a wealth of biomedical information, serving as valuable resources for the development of precision medicine systems. However, privacy concerns have resulted in limited access to high-quality and large-scale EHR data for researchers, impeding progress in methodological development. Recent research has delved into synthesizing realistic EHR data through generative modeling techniques, where a majority of proposed methods relied on generative adversarial networks (GAN) and their variants for EHR synthesis. Despite GAN-based methods attaining state-of-the-art performance in generating EHR data, these approaches are difficult to train and prone to mode collapse. Recently introduced in generative modeling, diffusion models have established cutting-edge performance in image generation, but their efficacy in EHR data synthesis remains largely unexplored. In this study, we investigate the potential of diffusion models for EHR data synthesis and introduce a novel method, \textsc{EHRDiff}. Through extensive experiments, \modelname establishes new state-of-the-art quality for synthetic EHR data, protecting private information in the meanwhile.
\end{abstract}

\section{Introduction}
Electronic health records (EHR) contain vast biomedical knowledge. 
EHR data may enable the development of state-of-the-art computational biomedical methods for dynamical treatment \citep{ASW2020expertsupervised}, differentiable diagnosis \citep{adpautodiag}, rare genetic disease identification \citep{Alsentzer2022.12.07.22283238}, etc. 
However, EHRs contain sensitive patients' private health information. Before being publicly accessible, real-world EHRs need to undergo de-identification \citep{Johnson2016MIMICIIIAF,Johnson2023MIMICIVAF}. 
The de-identification process uses automatic algorithms and requires tedious thorough human reviewing. Pending releasing approval can take months out of legal or ethical concerns \citep{10.1001/jama.282.15.1466}. 
Such circumstances limit the public release of rich EHR data, hence impeding the advancement of precision medicine methodologies. 
To mitigate the issue of limited publicly available EHR data, researchers alternatively explored generating synthetic EHR data \citep{pmlr-v68-choi17a,synthea}. 
Realistic synthetic EHR generation has recently become a research field of medical informatics. 

A line of work approached EHR data synthesis through generative modeling techniques, where they trained generative models on limited real EHR data to generate synthetic EHR data. 
Recent research developed variants of auto-encoders \citep{autoencoder,Biswal2020EVAGL} or generative adversarial networks (GAN) \citep{NIPS2014_gan,pmlr-v68-choi17a}. 
The majority of EHR data synthesis methodologies have relied on GAN \citep{pmlr-v68-choi17a,medbgan,EHRgan,Yan2020GeneratingEH}. 
Although GAN-based methods achieved state-of-the-art performance with respect to synthetic EHR quality and privacy preservation, they suffer from training instability and mode collapse \citep{che2017mode}. 
Previous research proposed different techniques to mitigate the problem, while as shown in our experiments, GAN-based methods still are prone to such problems, resulting in unsatisfactory synthetic data quality. 
This may raise concerns when developing real-world systems using synthetic EHR data from GAN-based methods.

Most recently, novel diffusion models \citep{pmlr-v37-sohl-dickstein15} in generative modeling have been proposed and have achieved cutting-edge generation performance in the field of vision \citep{ddpm,song2021scorebased}, audio \citep{kong2021diffwave}, or texts \citep{li2022diffusionlm,gong2023diffuseq,Yuan2022SeqDiffuSeqTD}. 
Many variants of diffusion models have surpassed the generation performance of GANs in sample quality and diversity. 
In general, starting from random noise features, diffusion models use a trained denoising distribution to gradually remove noise from the features and ultimately generate realistic synthetic features. 
The efficacy of diffusion models on realistic EHR synthesis is less studied compared to GANs. 
Considering the superior performance of diffusion models in other domains, our work explores the synthesizing performance of such techniques on EHR data. 
We introduce \modelname, a diffusion model-based EHR synthesizing model. 

Our work conducts comprehensive experiments using publicly available real EHR data and compares the effectiveness of \modelname against several other GAN-based EHR data synthesizing methods. 
We provide empirical evidence that \modelname is capable of generating synthetic EHR data with a high degree of quality. 
Additionally, our findings reveal that the synthetic EHR data produced by \modelname is of superior quality compared to those generated by GAN-based models, and it is more consistent with the distribution of real-world EHR data. 

Our research has two primary contributions: 
Firstly, we introduce the use of diffusion models to the realm of realistic EHR synthesis and propose a diffusion-based method called \modelname. 
Secondly, through extensive experimentation on publicly available EHR data, we demonstrate the superior quality of synthetic EHR data generated by \modelname in comparison to GAN-based EHR synthesizing methods. 
Furthermore, the synthetic EHR data generated by \modelname exhibits excellent correlation with real-world EHR data.

Our work is summarized as two following contributions: 
\begin{enumerate}
    \item We introduce diffusion models to the fields of EHR data synthesis and propose a diffusion-based method called \modelname. 
    \item Through extensive experiments on publicly available real EHR data, we empirically demonstrate the superior generation quality of \modelname over GAN-based EHR synthesis methods for various EHR feature formats, including categorical, continuous, and time-series features. 
    In the meanwhile, \modelname can safeguard private information in real training EHR.
\end{enumerate}

\section{Related Work}

\subsection{EHR data synthesis}

In the literature on EHR synthesis, researchers are usually concerned with the generation of discrete code features such as ICD codes rather than clinical narratives. Researchers have developed various methods to generate synthetic EHR data. Early work was usually disease-specific or covered a limited number of diseases. \citet{Buczak2010DatadrivenAF} developed a method that generates EHR including visit records, clinical activity, and laboratory results for 203 synthetic tularemia outbreak patients of tularemia. The features in synthetic EHR data are generated based on retrieving similar real-world EHR which is inflexible and prone to privacy leakage. \citet{synthea} developed a software named Synthea which generates synthetic EHRs with various patient information based on publicly available data. They build generation workflows based on biomedical knowledge and real-world feature statistics. Various models are aggregated for different feature synthesis. However, Synthea only covered the 20 most common conditions.

Recently, researchers mainly applied generative modeling methods for EHR synthesis \citep{Ghosheh2022ARO}. Medical GAN (medGAN) \citep{pmlr-v68-choi17a} introduced GAN to EHR synthesis. medGAN can generate synthetic EHR data with good quality and is free of tedious feature engineering. Following medGAN, various GAN-based methods are proposed, such as medBGAN \citep{medbgan}, EHRWGAN \citep{EHRgan}, CorGAN \citep{Torfi2020CorGANCC}, etc. These GAN-based methods advance synthetic EHR to higher quality. 
However, a common drawback of GAN-based methods is that these methods suffer from the mode collapse phenomenon which results in a circumstance where a GAN-based model is capable of generating only a few modes of real data distribution \citep{ThanhTung2018OnCF}.
To mitigate the problem, GAN methods for EHR generation rely on pre-trained auto-encoders to reduce the feature dimensions for training stability. However, inappropriate hyper-parameter choices and autoencoder pre-training will lead to sharp degradation of synthetic EHR quality or even failure to generate realistic data. There is also research that uses GAN-based models for conditional synthetic EHR generation to model the temporal structure of real EHR data \citep{Zhang2020SynTEGAF}. Since diffusion models are less studied in EHR synthesis, we focus on the unconditional generation of EHR and leave modeling conditional temporal structure with diffusion models to future work.  

Besides GAN-based methods, there also exists research that explores generating synthetic EHR data through variational auto-encoders \citep{Biswal2020EVAGL} or language models \citep{Wang2022PromptEHRCE}. 
Concurrently, MedDiff \citep{meddiff} is proposed and explores diffusion models for synthetic EHR generation, and they propose a new sampling technique without which the diffusion model fails to generate high-quality EHRs. \citet{ceritli2023synthesizing} directly apply TabDDPM \citep{kotelnikov2022tabddpm} to synthesizing EHR data. 

\subsection{Diffusion Models}

Diffusion models are formulated with forward and reverse processes. The forward process corrupts real-world data by gradually injecting noise, and harvesting training data with different noise levels for a denoising distribution, while the reverse process generates realistic data by removing noise using the denoising distribution. \citet{pmlr-v37-sohl-dickstein15} first proposed and provided theoretical support for diffusion models. Denoising diffusion probabilistic models (DDPM) \citep{ddpm} and noise conditional score networks (NCSN) \citep{song2021scorebased} have discovered the superior capability in image generation, and diffusion models have become a focused research direction since then. Recent research generalizes diffusion models to the synthesis of other data modalities and achieves excellent performance \citep{li2022diffusionlm,kong2021diffwave}. Our work is among the first research that introduces diffusion models to realistic EHR synthesis \citep{meddiff,ceritli2023synthesizing}.

\section{Method}

In this section, we give an introduction to the problem formulation of realistic EHR synthesis and the technical details of our proposed \modelname. 

\subsection{Problem Formulation}


Real-world EHR contains diverse patient information including demographics, physiological conditions, laboratory results, ICD codes, etc. 
Such information may be presented in various data formats (e.g., categorical, binary, continuous, etc). Features of different information may also have different scales, such as ages and heights. 
In this work, we use a vector $\boldsymbol{x}_0$ to represent the EHR, where $\boldsymbol{x}_0\in \mathcal{R}^ {|\mathcal{C}|}$ and $\mathcal{C}$ represents the dimension of EHR features of interest. 
We treat all features of different formats as real numbers ranging from $0$ to $1$. 
Following previous research \citep{pmlr-v68-choi17a,medbgan} which only focused on binary code data, we use $1$ standing for code occurrence and $0$ otherwise.
For continuous and categorical features, we normalize each feature within the range. 


\subsection{General Framework for Diffusion Model}

Diffusion models are characterized by forward and reverse Markov processes with latent variables where the forward process transforms the real samples to random Gaussian noise while the reverse process generates synthetic samples by gradual denoising the noise. As demonstrated by \citet{song2021scorebased}, the forward and reverse processes can be described by stochastic differential equations (SDE). The general SDE form for modeling the forward process follows:
\begin{align}
    \mathrm{d}\boldsymbol{x} = f(\boldsymbol{x}, t)\mathrm{d}t + g(t)\mathrm{d}\boldsymbol{w},
\end{align}
where $\boldsymbol{x}$ represents data points, $\boldsymbol{w}$ represents the standard Wiener process, $t$ is diffusion time, and ranges from $0$ to $T$. At $t=0$, $\boldsymbol{x}_0$ follows real data distribution while at $t=T$, $\boldsymbol{x}_T$ asymptotically follows a random Gaussian distribution. Functions $f$ and $g$ respectively define the sample corruption pattern and the level of injected noises. $f$ and $g$ together corrupt real-world samples to random noise. Based on the forward SDE, the SDE for the reverse process can be derived as follows:
\begin{align}
    \mathrm{d}\boldsymbol{x} = \left(f(\boldsymbol{x}, t)-g^2(t)\nabla_{\boldsymbol{x}}\log p_t(\boldsymbol{x})\right)\mathrm{d}t + g(t)\mathrm{d}\boldsymbol{w},
\end{align}
where $p_t(\boldsymbol{x})$ is the marginal density of $\boldsymbol{x}$ at time $t$, and $\nabla_{\boldsymbol{x}}\log p_t(\boldsymbol{x})$ is the score function, which indicates a vector field of which the direction is pointed to the high-density data area. 
With reparameterization proposed in \citet{Karras2022edm}, the reverse generation process can also be described with the probability flow ordinary differential equations (ODE) instead of SDEs \citep{song2021scorebased}:
\begin{align}
    &\mathrm{d}\boldsymbol{x}=-\dot{\sigma}(t)\sigma(t)\nabla_{x}\log p_t\big( \boldsymbol{x};\sigma(t) \big)\mathrm{d}t=-\dot{\sigma}(t)\sigma(t)\nabla_{x}\log p_{\sigma_t}\big( \boldsymbol{x} \big)\mathrm{d}t, \\
    &h(t)=\exp\left(\int_{0}^{t}f(\xi)\mathrm{d}\xi\right),\\
&\sigma(t)=\sigma_t=\sqrt{\int_{0}^t\frac{g(\xi)^2}{h(\xi)^2}\mathrm{d}\xi},
\label{odefin}
\end{align}
where $\dot{\sigma}(t)$ represents the derivative of $\sigma(t)$.
SDE and probability flow ODE indicate stochastic and deterministic generation processes respectively. To generate synthetic samples following reverse processes, it is required to learn a score function $s_\theta(\boldsymbol{x})$ parameterized by $\theta$ by score matching $\min_\theta \E_{p(\boldsymbol{x}_0)p_{\sigma_t}(\boldsymbol{x}|\boldsymbol{x}_0)}\left[\|s_\theta(\boldsymbol{x}) - \nabla_{\boldsymbol{x}}\log p_{\sigma_t}(\boldsymbol{x}|\boldsymbol{x}_0)\|^2_2\right]$.

\subsection{\modelname}

For the diffusion process of \modelname, we use $h(t)=1$ and $\sigma_t=t$ from previous research \citep{Karras2022edm}.
Therefore $\nabla_{\boldsymbol{x}}\log p_{\sigma_t}(\boldsymbol{x}|\boldsymbol{x_0}) = -\frac{\boldsymbol{x}-\boldsymbol{x}_0}{\sigma_t^2}$ and we reparameterize $s_\theta(\boldsymbol{x})=-\frac{\boldsymbol{x}-D_\theta(\boldsymbol{x},\sigma_t)}{\sigma_t^2}$, then the objective can be derived as:
\begin{align}
    &\min_\theta \E_{p(\boldsymbol{x}_0)p_{\sigma_t}(\boldsymbol{x}|\boldsymbol{x}_0)}\left[\left\|\frac{D_\theta(\boldsymbol{x},\sigma_t)-\boldsymbol{x}_0}{\sigma_t^2}\right\|^2_2\right],
\end{align}
and with further simplification of ignoring ${\sigma_t^2}$, the final objective becomes:
\begin{align} 
    \min_\theta \E_{p(\boldsymbol{x}_0)p_{\sigma_t}(\boldsymbol{x}|\boldsymbol{x}_0)}\left[\|D_\theta(\boldsymbol{x},\sigma_t)-\boldsymbol{x}_0\|^2_2\right]. \label{objective1}
\end{align}

Generally, $D_\theta(\boldsymbol{x},\sigma_t)$ can be modeled by neural networks, while such direct modeling may cause obstacles for optimization because the variance of $\boldsymbol{x}_t$ and the scale of $\sigma_t$ are diverse at different time step $t$. Therefore, we chose the pre-conditioning design of $D_\theta(\boldsymbol{x},\sigma_t)$ \citep{Karras2022edm} where $D_\theta(\boldsymbol{x},\sigma_t)$ is decomposed as:
\begin{align}
D_{\theta}(\boldsymbol{x};\sigma)=c_{\text{skip}}(\sigma)\boldsymbol{x}+c_{\text{out}}(\sigma)F_{\theta}(c_{\text{in}}(\sigma)\boldsymbol{x};c_{\text{noise}}(\sigma)).
\label{precond}
\end{align}
$F_{\theta}$ is modeled with neural networks and such designs of $c_{\text{in}}$ and $c_{\text{noise}}$ regulate the input to the network to be unit variance across different time step $t$, and $c_{\text{out}}$ and $c_{\text{skip}}$ together set the neural model prediction to be unit variance with minimized scale.
Therefore in \modelname, we chose $c_{\text{out}}=\sigma\sigma_{\text{data}}/\sqrt{\sigma^2+\sigma_{\text{data}}^2}$ and $c_{\text{skip}}(\sigma)=\sigma_{\text{data}}^2/(\sigma^2+\sigma_{\text{data}}^2)$. $c_{\text{noise}}(\sigma) = 0.25\ln{\sigma}$ which is designed empirically with the principle of constraining the input noise scale from varying immensely and $\ln(\sigma)\sim \mathcal{N}(P_{\text{mean}}, P_{\text{std}}^2)$ following \cite{Karras2022edm}, where $P_{\text{mean}}$ and $P_{\text{std}}$ are hyper-parameters to be set.

With the aforementioned formalization of $h(t)$, $\sigma(t)$ and the learned score function $s_\theta$, the generation process of \modelname can be expressed as the ODE:
\begin{align}
    \mathrm{d}\boldsymbol{x}=-t s_\theta(\boldsymbol{x})\mathrm{d}t. \label{odefin}
\end{align}
Solving the ODE numerically requires discretization of the time step $t$ and a proper design of noise level $\sigma_t$ along the solution trajectory. Therefore, following previous research \citep{Karras2022edm}, we set the maximum and minimum noise levels as $\sigma_{\text{max}}$ and $\sigma_{\text{min}}$, and use the following form of discretization:
\begin{align}
t_i=\sigma_{t_i}=\left((\sigma_{\text{max}})^{\frac{1}{\rho}}+\frac{i}{N-1}\left((\sigma_{\text{min}})^{\frac{1}{\rho}}-(\sigma_{\text{max}})^{\frac{1}{\rho}}\right)\right)^{\rho},
\end{align}
where $i$'s are integers and range from $0$ to $N$, $\sigma_{t_N} = 0, \sigma_{t_{N-1}} = \sigma_{\text{min}}$, and $\rho$ controls the schedules of discretized time step $t_i$ and trades off the discretized strides $t_i-t_{i-1}$ the larger value of which indicates a larger stride near $t_0$. 
In order to solve the ODE more precisely and generate synthetic EHR with higher quality, we use Heun's $2$nd order method, which adds a correction updating step for each $t_i$ and alleviates the truncation errors compared to the $1$st order Euler method. We leave the detailed sampling procedure to Appendix \ref{app:sampling_alg}.

\section{Experiments}

To demonstrate the effectiveness of our proposed \modelname, we conduct extensive experiments evaluating the quality of synthetic EHRs and the privacy concerns of the method. We also compare \modelname the several GAN-based realistic EHR synthesis methods to illustrate the performance of \modelname.

\subsection{Dataset}

Many previous research uses in-house EHR data which is not publicly available for method evaluation \citep{EHRgan,Yan2020GeneratingEH}. Such experiment designs set obstacles for later research to reproduce experiments. In this work, we use a publicly available EHR database, MIMIC-III, to evaluate \modelname. 

Deidentified and comprehensive clinical EHR data is integrated into MIMIC-III \citep{Johnson2016MIMICIIIAF}. The patients are admitted to the intensive care units of the Beth Israel Deaconess Medical Center in Boston. For each patient's EHR for one admission, we extract the diagnosis and procedure ICD-9 code and truncate the ICD-9 code to the first three digits. This preprocessing can reduce the long-tailed distribution of the ICD-9 code distribution and results in a 1,782 code set. Therefore, the EHR for each patient is formulated as a binary vector of 1,782 dimensions. The final extracted number of EHRs is 46,520 and we randomly select 41,868 for model training while the rest are held out for evaluation. 

Although most of the existing research focused on synthesizing discrete code features, real-world EHR data contains various data formats such as continuous test results values or time series of electrocardiograms (ECG). In this work, we extend the previous research and explore applying \modelname to the synthesis of EHR data other than binary codes. We use the following two datasets: CinC2012 Data and PTB-ECG Data. CinC2012 Data is a dataset for predicting the mortality of ICU patients and contains various feature formats such as categorical age, or continuous serum glucose values. PTB-ECG Data contains ECG signal data for heart disease diagnosis. Detailed introductions of both datasets are left in \ref{app:data}.
All the categorical features are converted into binary columns by one-hot encoding, and continuous features are normalized to values in the range of $[0.0, 1.0]$. We use sets A and B in CinC2023 Data as training and held-out testing sets respectively. The PTB-ECG Data is split with a ratio of 8:2 for training and held-out testing.

\subsection{Baselines}

To better demonstrate EHR synthesis performance, we compare \modelname to several strong baseline models as follows.

\paragraph{medGAN} \citep{pmlr-v68-choi17a} is the first work that introduces GAN to generating realistic synthetic EHR data. Considering the obstacle of directly using GAN to generate high-dimensional binary EHR vectors, medGAN alters to a low-dimensional dense space for generation by taking advantage of pre-trained auto-encoders. The model generates a dense EHR vector and then recovers a synthetic EHR with decoders.

\paragraph{medBGAN and medWGAN} \citep{medbgan} are two improved GAN models for realistic EHR synthesis. medGAN is based on the conventional GAN model for EHR synthesis, and such a model is prone to mode collapse where GAN models may fail to learn the distribution of real-world data. medBGAN and medWGAN integrate Boundary-seeking GAN (BGAN) \citep{devon2018boundary} and Wasserstein GAN (WGAN) \citep{wgan} respectively to improve the performance of medGAN and stabilize model training.

\paragraph{CorGAN}\citep{torfi2020corgan} is a novel work that utilizes convolutional neural networks (CNN) instead of multilayer perceptrons (MLP) to model EHR data. 
Specifically, they use CNN to model the autoencoder and the generative network. They empirically elucidate through experiments that CNN can perform better than the MLP in this task.

\paragraph{EMR-WGAN}\citep{zhang2020ensuring} is proposed to further refine the GAN models from several perspectives. To avoid model collapse, the authors take advantage of WGAN. The most prominent feature of EMR-WGAN is that it is directly trained on the discrete EHR data, while the previous research universally uses an autoencoder to first transform the raw EHR data into low-dimensional dense space. 
They utilize BatchNorm \citep{batchnorm} for the generator and LayerNorm \citep{layernorm} for the discriminator to improve performance.
As is shown in their experiments, these modifications significantly improve the performance of GAN.

\subsection{Evaluation Metric}
In our experiments, we evaluate the generative models' performance from two perspectives: utility and privacy \citep{Yan2022AMB}. Utility metrics evaluate the quality of synthetic EHRs and privacy metrics assess the risk of privacy breaches. In the following metrics, we generate and use the same number of synthetic EHR samples as the number of real training EHR samples.

\subsubsection{Utility Metrics}

We follow previous research for a set of utility metrics. The following metrics evaluate synthetic EHR quality from diverse perspectives.

\paragraph{Dimension-wise distribution} describes the feature-level resemblance between the synthetic data and the real data. 
The metric is widely used in previous research to investigate whether the generative model is capable of learning the high-dimensional distribution of real EHR data. 
For each code dimension, we calculate the empirical mean estimation for synthetic and real EHR data respectively. The mean estimation indicates the prevalence of the code. We visualize the dimension-wise distribution using scatter plots where both axes represent the prevalence of synthetic and real EHR respectively. 
Many codes have very low prevalence in real EHR data. The generation model may be prone to mode collapse and fail to generate the codes with low prevalence. Therefore, we count the number of codes that exist in the synthetic EHR samples and dub the quantity non-zero code columns.

\paragraph{Dimension-wise correlation} measures the difference between the feature correlation matrices of real and synthetic EHR data. The $i,j$ entry of correlation matrices calculates the Pearson correlation between the $i$th and $j$th features. For both the synthetic and real EHR data, we calculate first the correlation matrices, and then the averaged absolute differences between the correlation matrices. We name this metric the correlation matrix distance (CMD).

\paragraph{Dimension-wise prediction} evaluates whether generative models capture the inherent code feature relation by designing classification tasks. Specifically, we select one of the code features to be the classification target and use the rest of the features as predictors. To harvest a balanced target distribution, we sort the code features according to the entropy $H(p)$ of code prevalence $p$, where $H(p)=-p\log(p) -(1-p)\log(1-p)$. We select the top 30 code features according to entropies and form 30 individual classification tasks. For each task, we fit a classification model with logistic regression using real training and synthetic EHR data and assess the F1 score on the preset evaluation real EHR data.

\subsubsection{Privacy Metrics}

Generative modeling methods need real EHR data for training which raises privacy concerns among practitioners. Attackers may infer sensitive private information from trained models. Besides the utility of synthetic EHR data, we also evaluate existing models from a privacy protection perspective \citep{pmlr-v68-choi17a,EHRgan,Yan2022AMB}.

\paragraph{Attribute inference risk} describes the risk that sensitive private information of real EHR training data may be exposed based on the synthetic EHR data
It assumes a situation where the attackers already have several real EHR training samples with partially known features, and try to infer the rest features through synthetic data.  
Specifically, we assume that attackers first use the k-nearest neighbors method to find the top $k$ most similar synthetic EHRs to each real EHR based on the known code features, and then recover the rest of unknown code features by majority voting of $k$ similar synthetic EHRs. We set $k$ to 1 and use the most frequent 256 codes as the features known by the attackers. The metric is quantified by the prediction F1-score of the unknown code features.

\paragraph{Membership inference risk} evaluates the risk that given a set of real EHR samples, attackers may infer the samples used for training based on synthetic EHR data. We mix a subset of training real EHR data and held-out testing real EHR data to form an EHR set. For each EHR in this set, we calculate the minimum L2 distance with respect to the synthetic EHR data. The EHR whose distance is smaller than a preset threshold is predicted as the training EHR. We report the prediction F1 score to demonstrate the performance of each model under membership inference risk.

\begin{table}[t]
\caption{NZC represents Non-Zero code Columns, CMD represents Correlation Matrix Distance. $\downarrow$ and $\uparrow$ indicate the respectively lower and higher numbers for better results.}
\label{tab:utility}
\begin{center}
\begin{tabular}{ccccccc}
\toprule
        & NZC ($\uparrow$)& CMD ($\downarrow$)  \\ \midrule
medGAN   & 643$\pm$59.4 & 45.652$\pm$11.911   \\
medBGAN  & 898$\pm$36.8  & 63.186$\pm$16.359  \\
medWGAN  & 376$\pm$33.5   & 8.603$\pm$0.163    \\
CorGAN  & 753$\pm$125.1   & 10.997$\pm$0.420   \\
EMR-WGAN & 1060$\pm$29.6  & 8.173$\pm$0.274   \\
\modelname & \textbf{1770}$\pm$1.9&\textbf{7.769}$\pm$0.013  \\ \bottomrule
\end{tabular}
\end{center}
\end{table}

\begin{figure*}[t]
    \begin{center}
    \includegraphics[scale=0.65]{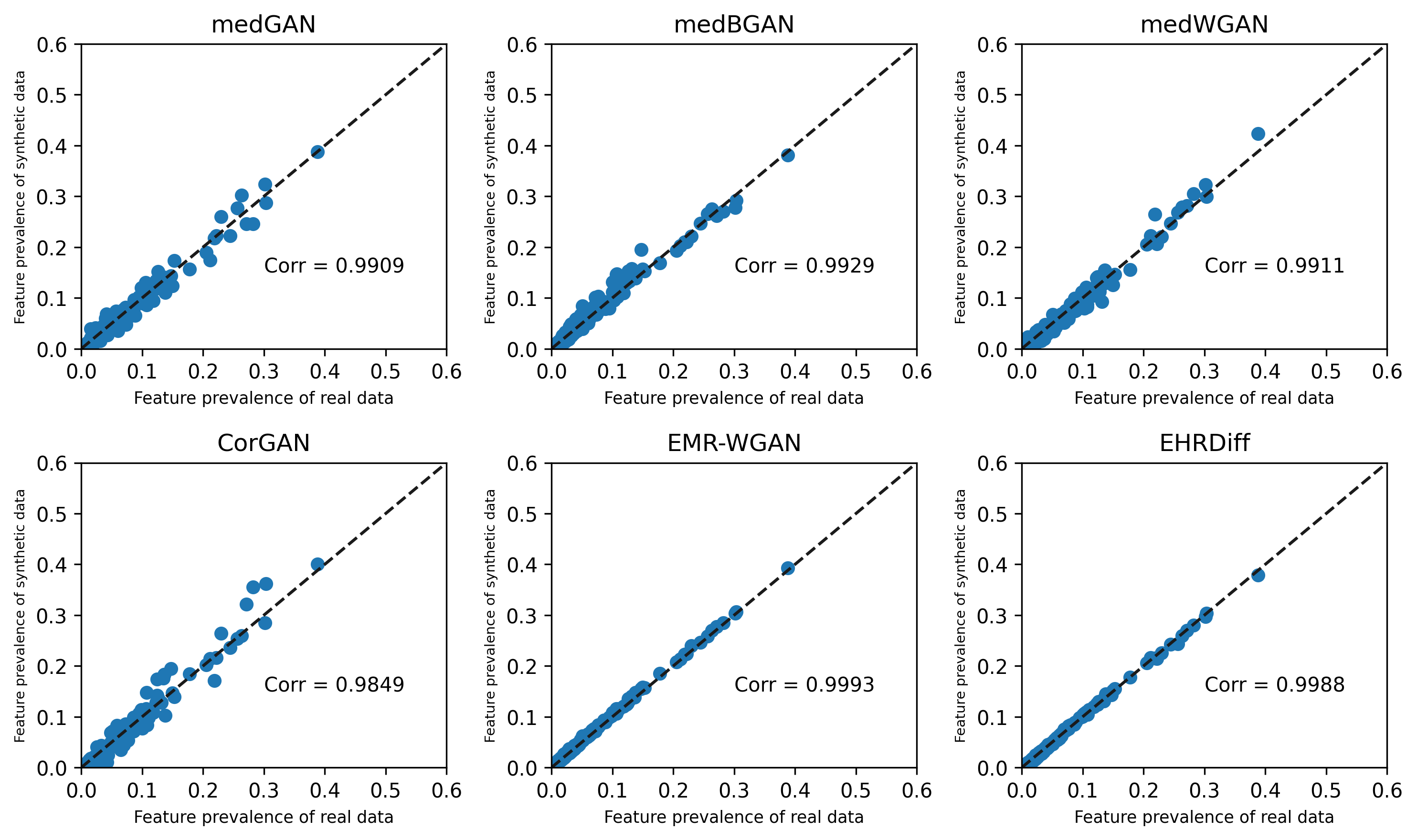}
    \end{center}
    \caption{The dimension-wise probability scatter plot of synthetic EHR data from different generative models against real EHR data. The diagonal lines represent the perfect match of code prevalence between synthetic and real EHR data. }
    \label{fig: dim-wise-scatter}
\end{figure*}

\begin{figure*}[t]
    \begin{center}
    \includegraphics[scale=0.65]{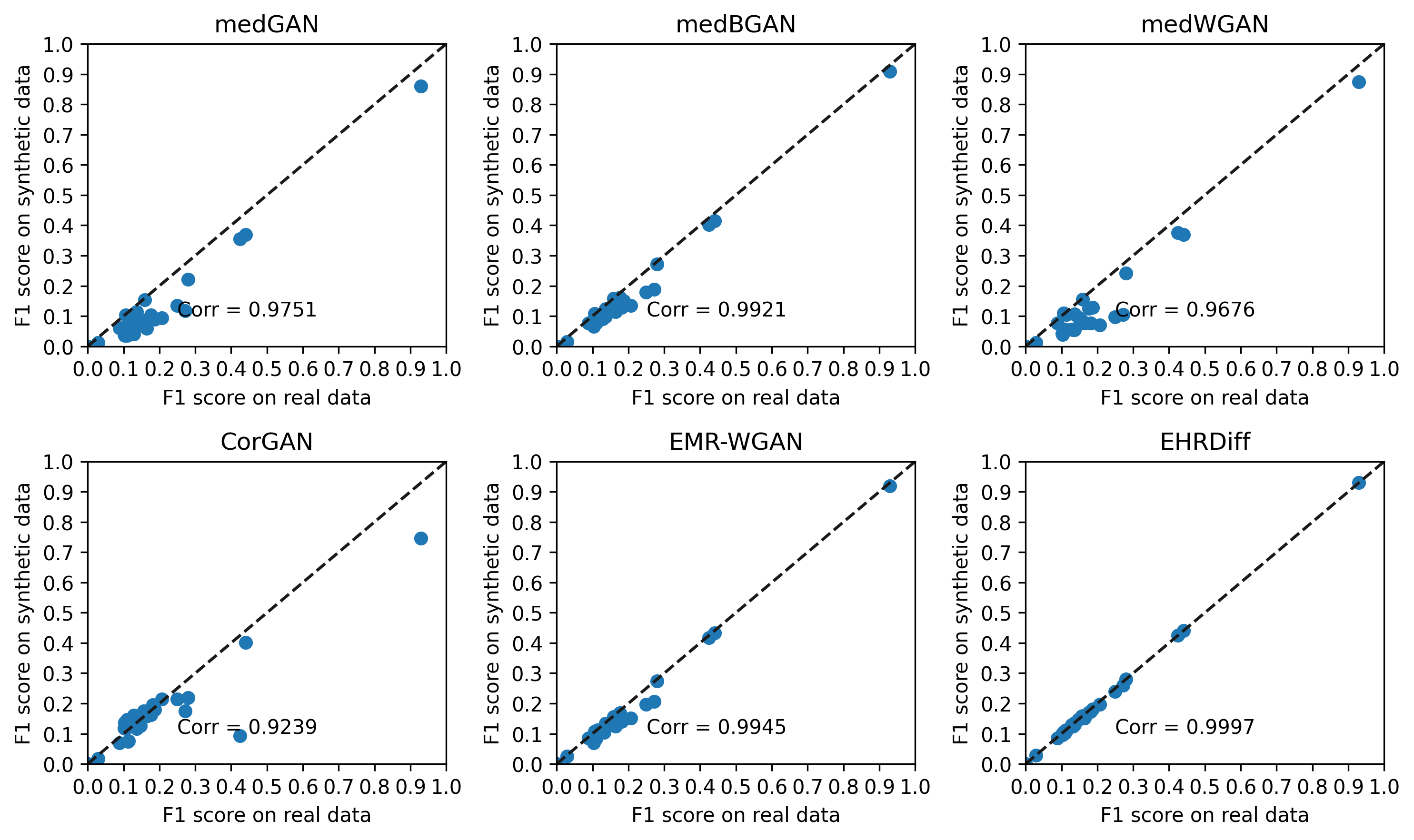}
    \end{center}
    \caption{The dimension-wise prediction scatter plot of synthetic EHR data from different generative models against real EHR data. The diagonal lines represent the perfect match of code prediction between synthetic and real EHR data. Each scatter represents a task.}
    \label{fig: dim-wise-predict}
\end{figure*}

\subsection{Implementation Detail}
In our experiments, for the diffusion noise schedule, we set $\sigma_{\text{min}}$ and $\sigma_{\text{max}}$ to be $0.02$ and $80$. $\rho$ is set to $7$ and the time step is discretized to $N=32$. $P_{mean}$ is set to $-1.2$ and $P_{std}$ is set to $1.2$ for noise distribution in the training process. For $F_\theta$ in Equation \ref{precond}, it is parameterized by an MLP with ReLU \citep{Nair2010RectifiedLU} activations and the hidden states are set to $[1024, 384, 384, 384, 1024]$. For the baseline methods, we follow the settings reported in their papers. The reported standard errors marked with $\pm$ are calculated under 5 different runs.

\subsection{Results on MIMIC}

\subsubsection{Utility Results}

Figure \ref{fig: dim-wise-scatter} depicts the dimension-wise prevalence distribution of synthetic EHR data against real data. The scatters from \modelname are distributed more closely to the diagonal dashed line compared to other baseline models, and \modelname and EMR-WGAN achieve near-perfect correlation. 
As shown in Table \ref{tab:utility}, \modelname outperforms all baseline methods in non-zero code column number (NZC) by large margins. This shows that GAN-based baselines all suffer from model collapse to different extents. The GAN-based method of best performance, EMR-WGAN, still fails to generate 722 code features with the same number of synthetic EHR samples as the real data. 
Although EMR-WGAN achieves a near-perfect correlation between real and synthetic code prevalence and is slightly better than \modelname, NZC demonstrates that the correlation can be biased by high prevalence features and overshadow the evaluation of low prevalence features. 
The results above demonstrate that \modelname can better capture the code feature prevalence of the real data than the GAN-based baselines, and is free from mode collapse. The synthetic EHR data by \modelname has better diversity than that by GAN-based methods.

From CMD results in Table \ref{tab:utility}, \modelname surpasses all baseline models. 
CMD results demonstrate that \modelname can better capture the inherent pair-wise relations between code features than GAN-based methods. 
The F1 score scatters in Figure \ref{fig: dim-wise-predict} of \modelname are closer to the diagonal lines and achieve the highest correlation value as compared to baselines. This means that training on synthetic EHR data by \modelname can lead to more similar performance to training on real data, and demonstrates that \modelname can better model complex interactions between code features than baselines.
It is indicated that synthetic EHR data by \modelname may have superior utility for training downstream models biomedical tasks.
We present more results on other utility metrics in \ref{app:results}.



\begin{table}[t]
\caption{The privacy assessment for each model. $\downarrow$ indicates the lower numbers for better results.}
\label{tab:privacy}
\begin{center}
\begin{tabular}{ccc}
\toprule
& Attribute Inference Risk ($\downarrow$)& Membership Inference Risk ($\downarrow$)\\\midrule
medGAN  & \textbf{0.0011}$\pm$0.0006 & 0.2941$\pm$0.0061 \\
medBGAN  & 0.0069$\pm$0.0025 & 0.2961$\pm$0.0047 \\
medWGAN  & 0.0066$\pm$0.0014 & 0.2928$\pm$0.0055 \\
CorGAN & 0.0033$\pm$0.0025 & \textbf{0.2413}$\pm$0.0061 \\
EMR-WGAN & 0.0259$\pm$0.0008 & 0.2943$\pm$0.0028 \\
\modelname & {0.0190}$\pm$0.0013&{0.2956}$\pm$0.0013\\ 
\bottomrule
\end{tabular}
\end{center}
\end{table}

\subsubsection{Privacy Results}\label{sec:privacy}

In Table \ref{tab:privacy}, we list the results against privacy attacks. In terms of attribute inference risk and membership inference risk, \modelname achieves intermediate results, while medGAN and CorGAN respectively achieve the best results on attribute inference risk and membership inference risk. However, as shown in utility results, the quality of synthetic EHR data by both models is far worse than \modelname. In an extreme circumstance where a generative model fails to fit the real EHR data distribution, the model may achieve perfect results on both privacy metrics, since attackers can not infer private information through synthetic data of bad quality. Therefore, there exists an implicit trade-off between utility and privacy. We suspect that medGAN and CorGAN can better safeguard privacy due to mediocre synthesis quality. When compared to EMR-WGAN which achieves the best synthesis quality among baselines, \modelname surpasses EMR-WGAN on attribute inference risk and achieves on-par results in terms of membership inference risk. To conclude, \modelname can well protect the sensitive private information of real EHR training data.

\begin{table}[t]
\caption{The AUC values on Cinc2012 Data and PTB-ECG Data.}
\label{tab:discussion}
\begin{center}
\begin{tabular}{ccccccc}
\toprule
         &  & Cinc2012 Data                               &  & PTB-ECG Data&  \\
Real     && 0.8479 &&0.9963 &   \\ \hline
medGAN   && 0.6176$\pm$0.0676  && 0.7550$\pm$0.0375 &    \\
medBGAN  && 0.5942$\pm$0.0763   && 0.7301$\pm$0.0215  &  \\
medWGAN  && 0.7012$\pm$0.0471  && 0.8071$\pm$0.0273 &     \\
CorGAN   && 0.6352$\pm$0.1259   && 0.4521$\pm$0.0709 &    \\
EMR-WGAN && 0.8010$\pm$0.0143   && 0.8011$\pm$0.0171  &  \\
EHRDiff  && \textbf{0.8405}$\pm$0.0024 &&  \textbf{0.9898}$\pm$0.0010 &   \\ \toprule
\end{tabular}
\end{center}
\end{table}

\subsection{Results on PTB-ECG and CinC2023}

Since both datasets are designed for classification, we inspect the utilities of synthesized data by evaluating the Area Under receiver operating characteristic Curve (AUC) of classifiers trained with synthetic data. We use LightGBM \citep{ke2017lightgbm} as classifiers and train on synthetic data of the same size as real training data. 

The results shown in Table \ref{tab:discussion} that classifiers trained by synthetic data from \modelname achieve the highest AUC values and are consistently better than GAN-based methods, reaching 0.8405 and 0.9898 on average on CinC2012 Data and PTB-ECG Data respectively. They also have on-par performance with classifiers trained by real data. The results show the great utility of \modelname generated EHR data, and the efficacy is consistently good across different EHR data feature formats. This demonstrates \modelname is practical in real-world scenarios and can approach EHR synthesis of diverse formats. The downstream biomedical models can benefit from training on synthetic EHR data by \modelname, potentially overcoming the obstacles of limited publicly available real EHR data.

\section{Discussion}

\begin{table}[t]
\caption{Ablation results of utility and privacy on different aspects of \modelname designs. VE and VP are short for variance exploding and variance preserving respectively. Pre-Cond represents pre-conditioning. }
    \label{tab:ablation}

\begin{center}
\begin{tabular}{l|ccc|cc}
\toprule
&\multicolumn{3}{c|}{Utility}&\multicolumn{2}{c}{Privacy}\\
&Corr ($\uparrow$) &NZC($\uparrow$)   &CMD ($\downarrow$)   &AIR ($\downarrow$)  &MIR ($\downarrow$) \\
\hline
\modelname    &0.9989$\pm$0.0001&1771.0$\pm$1.4&7.768$\pm$0.013& 0.0187$\pm$0.0010&0.2954$\pm$0.0014\\

\multicolumn{6}{l}{\textbf{on Diffusion Process}}\\
\hline
 w/ VE  &0.9876$\pm$0.0003& 1553.0$\pm$5.0 &8.057$\pm$0.022& 0.0130$\pm$0.0010 &0.2549$\pm$0.0023\\
 w/ VP  &0.9976$\pm$0.0001&1564.3$\pm$1.2& 8.002$\pm$0.013& 0.0143$\pm$0.0007& 0.3024$\pm$0.0007\\

\multicolumn{6}{l}{\textbf{on Sampling Method}}\\
\hline
 w/ 1st Euler& 0.9974$\pm$0.0001&1541.3$\pm$4.2&7.889$\pm$0.006& 0.0155$\pm$0.0003& 0.3111$\pm$0.0012\\

\multicolumn{6}{l}{\textbf{on Network Design}} \\
\hline
 w/o Pre-Cond&0.9655$\pm$0.0027& 432.0$\pm$26.8 &8.719$\pm$0.058& 0.0030$\pm$0.0009&0.3184$\pm$0.0018\\
\bottomrule
    \end{tabular}
    \end{center}
\end{table}

\subsection{Ablation Study}

In this section, we discuss and conduct ablation experiments on the effective designs of \modelname. The designs of \modelname include three major aspects: diffusion process, sampling method for inference, and neural network design for modeling denoising functions. We demonstrate the performance for each through the utility and privacy metrics on MIMIC-III. The results are shown in Table \ref{tab:ablation}. For the diffusion process, we compare the diffusion process of \modelname to the choice of variance exploding (VE) and variance preserving (VP) diffusion process. VE and VP are diffusion processes originally proposed in \citet{song2021scorebased}. We can see that \modelname outperforms both choices in terms of utility metrics while regarding the privacy concerns, \modelname performs a little inferior to VP and VE diffusion processes. As discussed in Section \ref{sec:privacy}, the privacy performance can be compromised to better utility performance. For the sampling process, Heun's 2nd order method outperforms Euler's 1st method in utility. A 2nd order sampling method can lead to a better step-wise estimation of each denoising step, and hence has better generation performance than 1st methods when using the same trained denoising function. For the network design, we can see that \modelname without pre-conditioning fails to generate high-quality EHR data as shown by the lowest NZC results of only 432. The possible explanation for this is that the binary code features may have diverse prevalence which leads to much difference in variance of each feature. Pre-conditioning sets the inputs and outputs of the neural networks to be unit variance and thus eases the modeling difficulty for the network in the denoising function.  

Overall, we have demonstrated the effectiveness of the design choices in \modelname regarding diffusion processes, sampling methods, and network designs. \modelname achieves the best performance regarding utility metrics, indicating a superior generation quality, while such designs in \modelname only compromise the privacy concerns marginally.

\begin{figure*}[t]
    \begin{center}
    \includegraphics[scale=0.46]{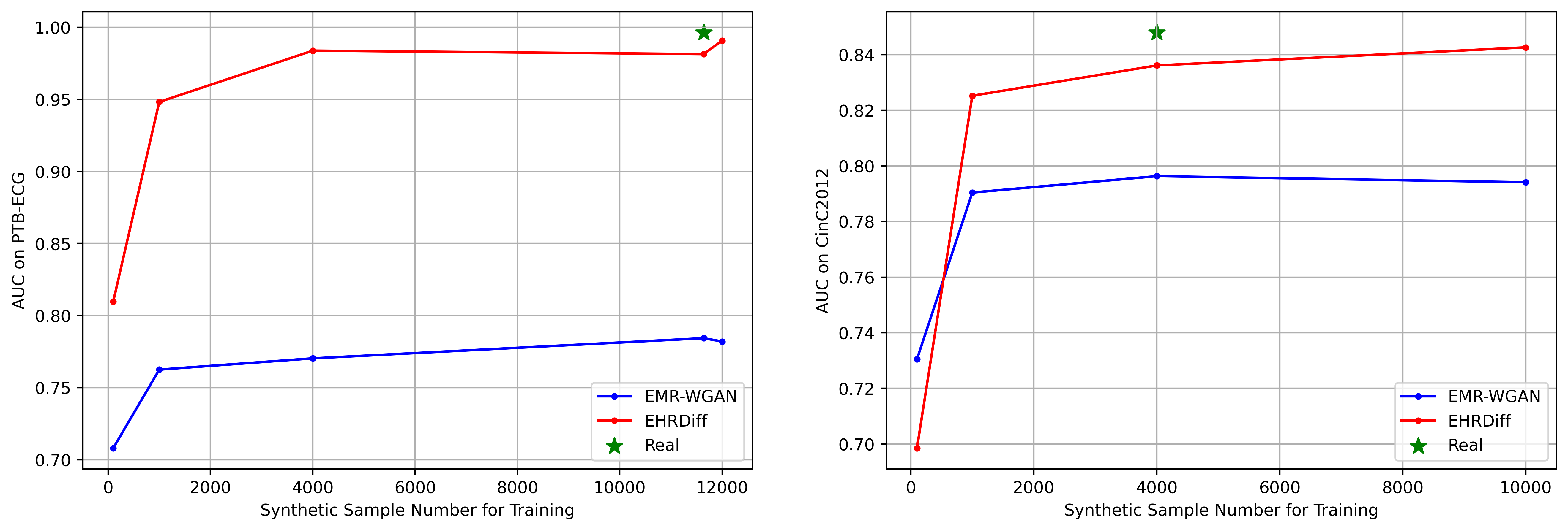}
    \end{center}
    \caption{The line plots for CinC2012 and PTB-ECG with different data scales. The green star represents the performance of the model trained on real data.}
    \label{fig: scale}
\end{figure*}

\subsection{Influence of Synthetic Data Scale}

Our main motivation is to use the synthetic data from \modelname to aid the downstream methodology development on EHR, emphasizing the limitation of scarce real-world EHR data. Therefore, we further demonstrate the influence of synthetic EHR data scales on training downstream models. We train classifiers of the same type on PTB-ECG and CinC2023 data with different scales of synthetic data from \modelname and the previous state-of-the-art EMR-WGAN for comparison. The results are shown in Figure \ref{fig: scale}.

As can be seen from the results, the downstream performance of models trained on synthetic data from \modelname improves as the size of synthesized data scales up. Models trained on synthetic data generated from \modelname outperformed those trained on data from EMR-WGAN across different sizes of synthetic training data for both CinC2012 and PTB-ECG, except on minimal data size of 100 on CinC2012. Notably, our results demonstrate that smaller sample sizes of synthetic EHR data from \modelname can lead to superior downstream model performance compared to larger sample sizes from EMR-WGAN. 

\section{Potential Social Impacts}

The generative models including our \modelname for synthesizing EHR data can potentially reveal their training data, resulting in unexpected patient privacy information leakage. 
Although in our experiments we conduct extensive experiments for privacy protection from attacking risks following existing research \citep{Yan2022AMB}, we also find the potential drawbacks of the privacy assessment metrics from our results and the extent to which applying our generative model in real-world application require further studying.

Data-driven generative models are prone to generating content with potential biases in training data. 
The EHR synthetic models may also suffer from such malicious tendencies of gender or demographic bias. Existing research on generative models has proposed methods to measure \citep{Teo2021MeasuringFI} or mitigate such problem \citep{Grover2019BiasCO,Grover2019FairGM,Teo2022FairGM}. 
Assessing and alleviating the bias of generative models for EHR synthesis remains an important research question.

\section{Conclusion}

In this work, we explore EHR data synthesis with diffusion models. We proposed \modelname, a diffusion-based model, for EHR data synthesis. Through comprehensive experiments on binary code EHR data, we empirically demonstrate the superior performance in generating high-quality synthetic EHR data from multiple evaluation perspectives, setting new state-of-the-art EHR synthesis methods. In the meanwhile, we also show \modelname can safeguard sensitive private information in real EHR training data. Furthermore, beyond binary code features in EHR data, the efficacy of \modelname consistently excels in continuous and time-series features. \modelname can help downstream biomedical methodology research overcome the obstacles of limited publicly available real EHR data.

\subsubsection*{Acknowledgment}
This work was supported by the Natural Science Foundation of China (Grant No. 12171270) and the Natural Science Foundation of Beijing Municipality (Grant No. Z190024).

\bibliography{main}

\begin{thebibliography}{50}
\providecommand{\natexlab}[1]{#1}
\providecommand{\url}[1]{\texttt{#1}}
\expandafter\ifx\csname urlstyle\endcsname\relax
  \providecommand{\doi}[1]{doi: #1}\else
  \providecommand{\doi}{doi: \begingroup \urlstyle{rm}\Url}\fi

\bibitem[Adler \& Lunz(2018)Adler and Lunz]{wgan}
Jonas Adler and Sebastian Lunz.
\newblock Banach wasserstein gan.
\newblock In S.~Bengio, H.~Wallach, H.~Larochelle, K.~Grauman, N.~Cesa-Bianchi,
  and R.~Garnett (eds.), \emph{Advances in Neural Information Processing
  Systems}, volume~31. Curran Associates, Inc., 2018.
\newblock URL
  \url{https://proceedings.neurips.cc/paper/2018/file/91d0dbfd38d950cb716c4dd26c5da08a-Paper.pdf}.

\bibitem[Alsentzer et~al.(2022)Alsentzer, Li, Kobren, Network, Kohane, and
  Zitnik]{Alsentzer2022.12.07.22283238}
Emily Alsentzer, Michelle~M. Li, Shilpa~N. Kobren, Undiagnosed~Diseases
  Network, Isaac~S. Kohane, and Marinka Zitnik.
\newblock Deep learning for diagnosing patients with rare genetic diseases.
\newblock \emph{medRxiv}, 2022.
\newblock \doi{10.1101/2022.12.07.22283238}.
\newblock URL
  \url{https://www.medrxiv.org/content/early/2022/12/13/2022.12.07.22283238}.

\bibitem[Ba et~al.(2016)Ba, Kiros, and Hinton]{layernorm}
Jimmy~Lei Ba, Jamie~Ryan Kiros, and Geoffrey~E. Hinton.
\newblock Layer normalization, 2016.
\newblock URL \url{https://arxiv.org/abs/1607.06450}.

\bibitem[Baowaly et~al.(2018)Baowaly, Lin, Liu, and Chen]{medbgan}
Mrinal~Kanti Baowaly, Chia-Ching Lin, Chao-Lin Liu, and Kuan-Ta Chen.
\newblock {Synthesizing electronic health records using improved generative
  adversarial networks}.
\newblock \emph{Journal of the American Medical Informatics Association},
  26\penalty0 (3):\penalty0 228--241, 12 2018.
\newblock ISSN 1527-974X.
\newblock \doi{10.1093/jamia/ocy142}.
\newblock URL \url{https://doi.org/10.1093/jamia/ocy142}.

\bibitem[Biswal et~al.(2020)Biswal, Ghosh, Duke, Malin, Stewart, and
  Sun]{Biswal2020EVAGL}
Siddharth Biswal, Soumya~Shubhra Ghosh, Jon~D. Duke, Bradley~A. Malin,
  Walter~F. Stewart, and Jimeng Sun.
\newblock Eva: Generating longitudinal electronic health records using
  conditional variational autoencoders.
\newblock \emph{ArXiv}, abs/2012.10020, 2020.

\bibitem[Bousseljot et~al.(1995)Bousseljot, Kreiseler, and
  Schnabel]{bousseljot1995nutzung}
Ralf Bousseljot, Dieter Kreiseler, and Allard Schnabel.
\newblock Nutzung der ekg-signaldatenbank cardiodat der ptb {\"u}ber das
  internet.
\newblock 1995.

\bibitem[Buczak et~al.(2010)Buczak, Babin, and Moniz]{Buczak2010DatadrivenAF}
Anna~L. Buczak, S.~Babin, and Linda~J. Moniz.
\newblock Data-driven approach for creating synthetic electronic medical
  records.
\newblock \emph{BMC Medical Informatics and Decision Making}, 10:\penalty0 59
  -- 59, 2010.

\bibitem[Ceritli et~al.(2023)Ceritli, Ghosheh, Chauhan, Zhu, Creagh, and
  Clifton]{ceritli2023synthesizing}
Taha Ceritli, Ghadeer~O. Ghosheh, Vinod~Kumar Chauhan, Tingting Zhu, Andrew~P.
  Creagh, and David~A. Clifton.
\newblock Synthesizing mixed-type electronic health records using diffusion
  models, 2023.

\bibitem[Che et~al.(2017)Che, Li, Jacob, Bengio, and Li]{che2017mode}
Tong Che, Yanran Li, Athul~Paul Jacob, Yoshua Bengio, and Wenjie Li.
\newblock Mode regularized generative adversarial networks, 2017.

\bibitem[Choi et~al.(2017)Choi, Biswal, Malin, Duke, Stewart, and
  Sun]{pmlr-v68-choi17a}
Edward Choi, Siddharth Biswal, Bradley Malin, Jon Duke, Walter~F. Stewart, and
  Jimeng Sun.
\newblock Generating multi-label discrete patient records using generative
  adversarial networks.
\newblock In Finale Doshi-Velez, Jim Fackler, David Kale, Rajesh Ranganath,
  Byron Wallace, and Jenna Wiens (eds.), \emph{Proceedings of the 2nd Machine
  Learning for Healthcare Conference}, volume~68 of \emph{Proceedings of
  Machine Learning Research}, pp.\  286--305. PMLR, 18--19 Aug 2017.
\newblock URL \url{https://proceedings.mlr.press/v68/choi17a.html}.

\bibitem[Ghosheh et~al.(2022)Ghosheh, Li, and Zhu]{Ghosheh2022ARO}
Ghadeer~O. Ghosheh, Jin Li, and Tingting Zhu.
\newblock A review of generative adversarial networks for electronic health
  records: applications, evaluation measures and data sources.
\newblock \emph{ArXiv}, abs/2203.07018, 2022.

\bibitem[Gong et~al.(2023)Gong, Li, Feng, Wu, and Kong]{gong2023diffuseq}
Shansan Gong, Mukai Li, Jiangtao Feng, Zhiyong Wu, and Lingpeng Kong.
\newblock Diffuseq: Sequence to sequence text generation with diffusion models,
  2023.

\bibitem[Goodfellow et~al.(2014)Goodfellow, Pouget-Abadie, Mirza, Xu,
  Warde-Farley, Ozair, Courville, and Bengio]{NIPS2014_gan}
Ian Goodfellow, Jean Pouget-Abadie, Mehdi Mirza, Bing Xu, David Warde-Farley,
  Sherjil Ozair, Aaron Courville, and Yoshua Bengio.
\newblock Generative adversarial nets.
\newblock In Z.~Ghahramani, M.~Welling, C.~Cortes, N.~Lawrence, and K.Q.
  Weinberger (eds.), \emph{Advances in Neural Information Processing Systems},
  volume~27. Curran Associates, Inc., 2014.
\newblock URL
  \url{https://proceedings.neurips.cc/paper/2014/file/5ca3e9b122f61f8f06494c97b1afccf3-Paper.pdf}.

\bibitem[Grover et~al.(2019{\natexlab{a}})Grover, Choi, Shu, and
  Ermon]{Grover2019FairGM}
Aditya Grover, Kristy Choi, Rui Shu, and Stefano Ermon.
\newblock Fair generative modeling via weak supervision.
\newblock In \emph{International Conference on Machine Learning},
  2019{\natexlab{a}}.
\newblock URL \url{https://api.semanticscholar.org/CorpusID:204904806}.

\bibitem[Grover et~al.(2019{\natexlab{b}})Grover, Song, Agarwal, Tran, Kapoor,
  Horvitz, and Ermon]{Grover2019BiasCO}
Aditya Grover, Jiaming Song, Alekh Agarwal, Kenneth Tran, Ashish Kapoor, Eric
  Horvitz, and Stefano Ermon.
\newblock Bias correction of learned generative models using likelihood-free
  importance weighting.
\newblock \emph{ArXiv}, abs/1906.09531, 2019{\natexlab{b}}.
\newblock URL \url{https://api.semanticscholar.org/CorpusID:195345228}.

\bibitem[He et~al.(2023)He, Zhao, Xi, and Ho]{meddiff}
Huan He, Shifan Zhao, Yuanzhe Xi, and Joyce~C Ho.
\newblock Meddiff: Generating electronic health records using accelerated
  denoising diffusion model, 2023.
\newblock URL \url{https://arxiv.org/abs/2302.04355}.

\bibitem[Hjelm et~al.(2018)Hjelm, Jacob, Trischler, Che, Cho, and
  Bengio]{devon2018boundary}
R~Devon Hjelm, Athul~Paul Jacob, Adam Trischler, Gerry Che, Kyunghyun Cho, and
  Yoshua Bengio.
\newblock Boundary seeking {GAN}s.
\newblock In \emph{International Conference on Learning Representations}, 2018.
\newblock URL \url{https://openreview.net/forum?id=rkTS8lZAb}.

\bibitem[Ho et~al.(2020)Ho, Jain, and Abbeel]{ddpm}
Jonathan Ho, Ajay Jain, and Pieter Abbeel.
\newblock Denoising diffusion probabilistic models.
\newblock In H.~Larochelle, M.~Ranzato, R.~Hadsell, M.F. Balcan, and H.~Lin
  (eds.), \emph{Advances in Neural Information Processing Systems}, volume~33,
  pp.\  6840--6851. Curran Associates, Inc., 2020.
\newblock URL
  \url{https://proceedings.neurips.cc/paper_files/paper/2020/file/4c5bcfec8584af0d967f1ab10179ca4b-Paper.pdf}.

\bibitem[Hodge et~al.(1999)Hodge, Gostin, and
  Jacobson]{10.1001/jama.282.15.1466}
James~G. Hodge, Jr, Lawrence~O. Gostin, and Peter~D. Jacobson.
\newblock {Legal Issues Concerning Electronic Health InformationPrivacy,
  Quality, and Liability}.
\newblock \emph{JAMA}, 282\penalty0 (15):\penalty0 1466--1471, 10 1999.
\newblock ISSN 0098-7484.
\newblock \doi{10.1001/jama.282.15.1466}.
\newblock URL \url{https://doi.org/10.1001/jama.282.15.1466}.

\bibitem[Ioffe \& Szegedy(2015)Ioffe and Szegedy]{batchnorm}
Sergey Ioffe and Christian Szegedy.
\newblock Batch normalization: Accelerating deep network training by reducing
  internal covariate shift.
\newblock In Francis Bach and David Blei (eds.), \emph{Proceedings of the 32nd
  International Conference on Machine Learning}, volume~37 of \emph{Proceedings
  of Machine Learning Research}, pp.\  448--456, Lille, France, 07--09 Jul
  2015. PMLR.
\newblock URL \url{https://proceedings.mlr.press/v37/ioffe15.html}.

\bibitem[Johnson et~al.(2016)Johnson, Pollard, Shen, wei H.~Lehman, Feng,
  Ghassemi, Moody, Szolovits, Celi, and Mark]{Johnson2016MIMICIIIAF}
Alistair E.~W. Johnson, Tom~J. Pollard, Lu~Shen, Li~wei H.~Lehman, Mengling
  Feng, Mohammad~Mahdi Ghassemi, Benjamin Moody, Peter Szolovits, Leo~Anthony
  Celi, and Roger~G. Mark.
\newblock Mimic-iii, a freely accessible critical care database.
\newblock \emph{Scientific Data}, 3, 2016.

\bibitem[Johnson et~al.(2023)Johnson, Bulgarelli, Shen, Gayles, Shammout,
  Horng, Pollard, Moody, Gow, wei H.~Lehman, Celi, and
  Mark]{Johnson2023MIMICIVAF}
Alistair E.~W. Johnson, Lucas Bulgarelli, Lu~Shen, Alvin Gayles, Ayad Shammout,
  Steven Horng, Tom~J. Pollard, Benjamin Moody, Brian Gow, Li~wei H.~Lehman,
  Leo~Anthony Celi, and Roger~G. Mark.
\newblock Mimic-iv, a freely accessible electronic health record dataset.
\newblock \emph{Scientific Data}, 10, 2023.

\bibitem[Johnson et~al.(2012)Johnson, Dunkley, Mayaud, Tsanas, Kramer, and
  Clifford]{johnson2012patient}
Alistair~EW Johnson, Nic Dunkley, Louis Mayaud, Athanasios Tsanas, Andrew~A
  Kramer, and Gari~D Clifford.
\newblock Patient specific predictions in the intensive care unit using a
  bayesian ensemble.
\newblock In \emph{Computing in Cardiology (CinC), 2012}, pp.\  249--252. IEEE,
  2012.

\bibitem[Kachuee et~al.(2018)Kachuee, Fazeli, and Sarrafzadeh]{kachuee2018ecg}
Mohammad Kachuee, Shayan Fazeli, and Majid Sarrafzadeh.
\newblock Ecg heartbeat classification: A deep transferable representation.
\newblock In \emph{2018 IEEE international conference on healthcare informatics
  (ICHI)}, pp.\  443--444. IEEE, 2018.

\bibitem[Karras et~al.(2022)Karras, Aittala, Aila, and Laine]{Karras2022edm}
Tero Karras, Miika Aittala, Timo Aila, and Samuli Laine.
\newblock Elucidating the design space of diffusion-based generative models.
\newblock In \emph{Proc. NeurIPS}, 2022.

\bibitem[Ke et~al.(2017)Ke, Meng, Finley, Wang, Chen, Ma, Ye, and
  Liu]{ke2017lightgbm}
Guolin Ke, Qi~Meng, Thomas Finley, Taifeng Wang, Wei Chen, Weidong Ma, Qiwei
  Ye, and Tie-Yan Liu.
\newblock Lightgbm: A highly efficient gradient boosting decision tree.
\newblock \emph{Advances in neural information processing systems}, 30, 2017.

\bibitem[Kong et~al.(2021)Kong, Ping, Huang, Zhao, and
  Catanzaro]{kong2021diffwave}
Zhifeng Kong, Wei Ping, Jiaji Huang, Kexin Zhao, and Bryan Catanzaro.
\newblock Diffwave: A versatile diffusion model for audio synthesis.
\newblock In \emph{International Conference on Learning Representations}, 2021.
\newblock URL \url{https://openreview.net/forum?id=a-xFK8Ymz5J}.

\bibitem[Kotelnikov et~al.(2022)Kotelnikov, Baranchuk, Rubachev, and
  Babenko]{kotelnikov2022tabddpm}
Akim Kotelnikov, Dmitry Baranchuk, Ivan Rubachev, and Artem Babenko.
\newblock Tabddpm: Modelling tabular data with diffusion models, 2022.

\bibitem[Li et~al.(2022)Li, Thickstun, Gulrajani, Liang, and
  Hashimoto]{li2022diffusionlm}
Xiang~Lisa Li, John Thickstun, Ishaan Gulrajani, Percy Liang, and Tatsunori
  Hashimoto.
\newblock Diffusion-{LM} improves controllable text generation.
\newblock In Alice~H. Oh, Alekh Agarwal, Danielle Belgrave, and Kyunghyun Cho
  (eds.), \emph{Advances in Neural Information Processing Systems}, 2022.
\newblock URL \url{https://openreview.net/forum?id=3s9IrEsjLyk}.

\bibitem[Nair \& Hinton(2010)Nair and Hinton]{Nair2010RectifiedLU}
Vinod Nair and Geoffrey~E. Hinton.
\newblock Rectified linear units improve restricted boltzmann machines.
\newblock In \emph{International Conference on Machine Learning}, 2010.
\newblock URL \url{https://api.semanticscholar.org/CorpusID:15539264}.

\bibitem[Silva et~al.(2012)Silva, Moody, Scott, Celi, and
  Mark]{silva2012predicting}
Ikaro Silva, George Moody, Daniel~J Scott, Leo~A Celi, and Roger~G Mark.
\newblock Predicting in-hospital mortality of icu patients: The
  physionet/computing in cardiology challenge 2012.
\newblock \emph{Computing in cardiology}, 39:\penalty0 245, 2012.

\bibitem[Sohl-Dickstein et~al.(2015)Sohl-Dickstein, Weiss, Maheswaranathan, and
  Ganguli]{pmlr-v37-sohl-dickstein15}
Jascha Sohl-Dickstein, Eric Weiss, Niru Maheswaranathan, and Surya Ganguli.
\newblock Deep unsupervised learning using nonequilibrium thermodynamics.
\newblock In Francis Bach and David Blei (eds.), \emph{Proceedings of the 32nd
  International Conference on Machine Learning}, volume~37 of \emph{Proceedings
  of Machine Learning Research}, pp.\  2256--2265, Lille, France, 07--09 Jul
  2015. PMLR.
\newblock URL \url{https://proceedings.mlr.press/v37/sohl-dickstein15.html}.

\bibitem[Sonabend et~al.(2020)Sonabend, Lu, Celi, Cai, and
  Szolovits]{ASW2020expertsupervised}
Aaron Sonabend, Junwei Lu, Leo~Anthony Celi, Tianxi Cai, and Peter Szolovits.
\newblock Expert-supervised reinforcement learning for offline policy learning
  and evaluation.
\newblock In \emph{Advances in Neural Information Processing Systems},
  volume~33, pp.\  18967--18977, 2020.
\newblock URL
  \url{https://proceedings.neurips.cc/paper/2020/file/daf642455364613e2120c636b5a1f9c7-Paper.pdf}.

\bibitem[Song et~al.(2021)Song, Sohl-Dickstein, Kingma, Kumar, Ermon, and
  Poole]{song2021scorebased}
Yang Song, Jascha Sohl-Dickstein, Diederik~P Kingma, Abhishek Kumar, Stefano
  Ermon, and Ben Poole.
\newblock Score-based generative modeling through stochastic differential
  equations.
\newblock In \emph{International Conference on Learning Representations}, 2021.
\newblock URL \url{https://openreview.net/forum?id=PxTIG12RRHS}.

\bibitem[Teo \& Cheung(2021)Teo and Cheung]{Teo2021MeasuringFI}
Christopher T.~H. Teo and Ngai-Man Cheung.
\newblock Measuring fairness in generative models.
\newblock \emph{ArXiv}, abs/2107.07754, 2021.
\newblock URL \url{https://api.semanticscholar.org/CorpusID:236034107}.

\bibitem[Teo et~al.(2022)Teo, Abdollahzadeh, and Cheung]{Teo2022FairGM}
Christopher T.~H. Teo, Milad Abdollahzadeh, and Ngai-Man Cheung.
\newblock Fair generative models via transfer learning.
\newblock \emph{ArXiv}, abs/2212.00926, 2022.
\newblock URL \url{https://api.semanticscholar.org/CorpusID:254221195}.

\bibitem[Thanh-Tung et~al.(2018)Thanh-Tung, Tran, and
  Venkatesh]{ThanhTung2018OnCF}
Hoang Thanh-Tung, T.~Tran, and Svetha Venkatesh.
\newblock On catastrophic forgetting and mode collapse in generative
  adversarial networks.
\newblock \emph{ArXiv}, abs/1807.04015, 2018.

\bibitem[Torfi \& Fox(2020{\natexlab{a}})Torfi and Fox]{Torfi2020CorGANCC}
Amirsina Torfi and Edward~A. Fox.
\newblock Corgan: Correlation-capturing convolutional generative adversarial
  networks for generating synthetic healthcare records.
\newblock In \emph{The Florida AI Research Society}, 2020{\natexlab{a}}.

\bibitem[Torfi \& Fox(2020{\natexlab{b}})Torfi and Fox]{torfi2020corgan}
Amirsina Torfi and Edward~A Fox.
\newblock Corgan: Correlation-capturing convolutional generative adversarial
  networks for generating synthetic healthcare records.
\newblock \emph{arXiv preprint arXiv:2001.09346}, 2020{\natexlab{b}}.

\bibitem[Vincent et~al.(2008)Vincent, Larochelle, Bengio, and
  Manzagol]{autoencoder}
Pascal Vincent, Hugo Larochelle, Yoshua Bengio, and Pierre-Antoine Manzagol.
\newblock Extracting and composing robust features with denoising autoencoders.
\newblock In \emph{Proceedings of the 25th International Conference on Machine
  Learning}, ICML '08, pp.\  1096–1103, New York, NY, USA, 2008. Association
  for Computing Machinery.
\newblock ISBN 9781605582054.
\newblock \doi{10.1145/1390156.1390294}.
\newblock URL \url{https://doi.org/10.1145/1390156.1390294}.

\bibitem[Walonoski et~al.(2017)Walonoski, Kramer, Nichols, Quina, Moesel, Hall,
  Duffett, Dube, Gallagher, and McLachlan]{synthea}
Jason Walonoski, Mark Kramer, Joseph Nichols, Andre Quina, Chris Moesel, Dylan
  Hall, Carlton Duffett, Kudakwashe Dube, Thomas Gallagher, and Scott
  McLachlan.
\newblock {Synthea: An approach, method, and software mechanism for generating
  synthetic patients and the synthetic electronic health care record}.
\newblock \emph{Journal of the American Medical Informatics Association},
  25\penalty0 (3):\penalty0 230--238, 08 2017.
\newblock ISSN 1527-974X.
\newblock \doi{10.1093/jamia/ocx079}.
\newblock URL \url{https://doi.org/10.1093/jamia/ocx079}.

\bibitem[Wang \& Sun(2022)Wang and Sun]{Wang2022PromptEHRCE}
Zifeng Wang and Jimeng Sun.
\newblock Promptehr: Conditional electronic healthcare records generation with
  prompt learning.
\newblock In \emph{Conference on Empirical Methods in Natural Language
  Processing}, 2022.

\bibitem[Yan et~al.(2020)Yan, Zhang, Nyemba, and Malin]{Yan2020GeneratingEH}
Chao Yan, Ziqi Zhang, Steve Nyemba, and Bradley~A. Malin.
\newblock Generating electronic health records with multiple data types and
  constraints.
\newblock \emph{AMIA Annual Symposium proceedings. AMIA Symposium},
  2020:\penalty0 1335--1344, 2020.

\bibitem[Yan et~al.(2022)Yan, Yan, Wan, Zhang, Omberg, Guinney, Mooney, and
  Malin]{Yan2022AMB}
Chao Yan, Yao Yan, Zhiyu Wan, Ziqi Zhang, Larsson Omberg, Justin Guinney,
  Sean~D. Mooney, and Bradley~A. Malin.
\newblock A multifaceted benchmarking of synthetic electronic health record
  generation models.
\newblock \emph{Nature Communications}, 13, 2022.

\bibitem[Yuan \& Yang(2019)Yuan and Yang]{yuan2019research}
Chunhui Yuan and Haitao Yang.
\newblock Research on k-value selection method of k-means clustering algorithm.
\newblock \emph{J}, 2\penalty0 (2):\penalty0 226--235, 2019.

\bibitem[Yuan \& Yu(2021)Yuan and Yu]{adpautodiag}
Hongyi Yuan and Sheng Yu.
\newblock Efficient symptom inquiring and diagnosis via adaptive alignment of
  reinforcement learning and classification.
\newblock \emph{CoRR}, abs/2112.00733, 2021.
\newblock URL \url{https://arxiv.org/abs/2112.00733}.

\bibitem[Yuan et~al.(2022)Yuan, Yuan, Tan, Huang, and
  Huang]{Yuan2022SeqDiffuSeqTD}
Hongyi Yuan, Zheng Yuan, Chuanqi Tan, Fei Huang, and Songfang Huang.
\newblock Seqdiffuseq: Text diffusion with encoder-decoder transformers.
\newblock \emph{ArXiv}, abs/2212.10325, 2022.

\bibitem[Zhang et~al.(2019)Zhang, Yan, Mesa, Sun, and Malin]{EHRgan}
Ziqi Zhang, Chao Yan, Diego~A Mesa, Jimeng Sun, and Bradley~A Malin.
\newblock {Ensuring electronic medical record simulation through better
  training, modeling, and evaluation}.
\newblock \emph{Journal of the American Medical Informatics Association},
  27\penalty0 (1):\penalty0 99--108, 10 2019.
\newblock ISSN 1527-974X.
\newblock \doi{10.1093/jamia/ocz161}.
\newblock URL \url{https://doi.org/10.1093/jamia/ocz161}.

\bibitem[Zhang et~al.(2020{\natexlab{a}})Zhang, Yan, Lasko, Sun, and
  Malin]{Zhang2020SynTEGAF}
Ziqi Zhang, Chao Yan, Thomas~A. Lasko, Jimeng Sun, and Bradley~A. Malin.
\newblock Synteg: a framework for temporal structured electronic health data
  simulation.
\newblock \emph{Journal of the American Medical Informatics Association :
  JAMIA}, 2020{\natexlab{a}}.

\bibitem[Zhang et~al.(2020{\natexlab{b}})Zhang, Yan, Mesa, Sun, and
  Malin]{zhang2020ensuring}
Ziqi Zhang, Chao Yan, Diego~A Mesa, Jimeng Sun, and Bradley~A Malin.
\newblock Ensuring electronic medical record simulation through better
  training, modeling, and evaluation.
\newblock \emph{Journal of the American Medical Informatics Association},
  27\penalty0 (1):\penalty0 99--108, 2020{\natexlab{b}}.

\end{thebibliography}
\bibliographystyle{tmlr}
\newpage
\appendix

\section{Sampling Algorithm}\label{app:sampling_alg}

\begin{algorithm}[t] 
\caption{Heun's $2$nd Method for Sampling} 
\label{alg:Framwork} 
\begin{algorithmic}[1] 
\ENSURE Time Step $t_i$ and noise level $\sigma_{t_i}$
\STATE Using Equation \ref{odefin}, calculate the derivative $\boldsymbol{g}_{t_i}=\mathrm{d}\boldsymbol{x}/\mathrm{d}t$:\\
$\boldsymbol{g}_{t_i}=t_i^{-1}\boldsymbol{x}_{t_i}-t_i^{-1}D(\boldsymbol{x}_{t_i};\sigma_{t_i})$,
\STATE Get intermediate $\Tilde{\boldsymbol{x}}_{t_{i+1}}$ by taking Euler step: \\
$\Tilde{\boldsymbol{x}}_{t_{i+1}}=\boldsymbol{g}_{t_i}(t_{i+1}-t_i) + \boldsymbol{x}_{t_i}$,
\STATE Calculate the gradient correction $\Tilde{\boldsymbol{g}}_{t_i}$: \\
$\Tilde{\boldsymbol{g}}_{t_i}=t_{i+1}^{-1}\Tilde{\boldsymbol{x}}_{t_{i+1}}-t_{i+1}^{-1}D(\Tilde{\boldsymbol{x}}_{t_{i+1}};\sigma_{t_{i+1}})$,
\STATE Get next time step sample $\boldsymbol{x}_{t_{i+1}}$: \\
$\boldsymbol{x}_{i+1}=\boldsymbol{x}_i+(t_{i+1}-t_i)(\frac{1}{2}\boldsymbol{g}_i+\frac{1}{2}\Tilde{\boldsymbol{g}}_{i+1})$
\RETURN $\boldsymbol{x}_{t_{i+1}}$
\end{algorithmic}
\end{algorithm}

\section{Additional Results}
\label{app:results}

\begin{table}[t]
\caption{MCAD represents Medical Concept Abundance Distance. $\downarrow$ and $\uparrow$ indicate the respectively lower and higher numbers for better results.}
\label{tabapp:utility}
\begin{center}

\begin{tabular}{ccccccc}
\toprule
          & Latent Distance ($\downarrow$) & MCAD ($\downarrow$)\\ \midrule
medGAN   & -4.300$\pm$0.009  & 0.257$\pm$0.014 \\
medBGAN   & -5.407$\pm$2.452 & 0.123$\pm$0.010  \\
medWGAN & -12.968$\pm$1.600 & 0.076$\pm$0.024  \\
CorGAN  & -9.917$\pm$0.771 &  0.129$\pm$0.053 \\
EMR-WGAN  & -14.437$\pm$0.549 &  0.101$\pm$0.008 \\
\modelname   & -13.560$\pm$0.211 & 0.076$\pm$0.002   \\ \bottomrule
\end{tabular}
\end{center}
\end{table}

\begin{figure*}[t]
    \begin{center}
    \includegraphics[scale=0.6]{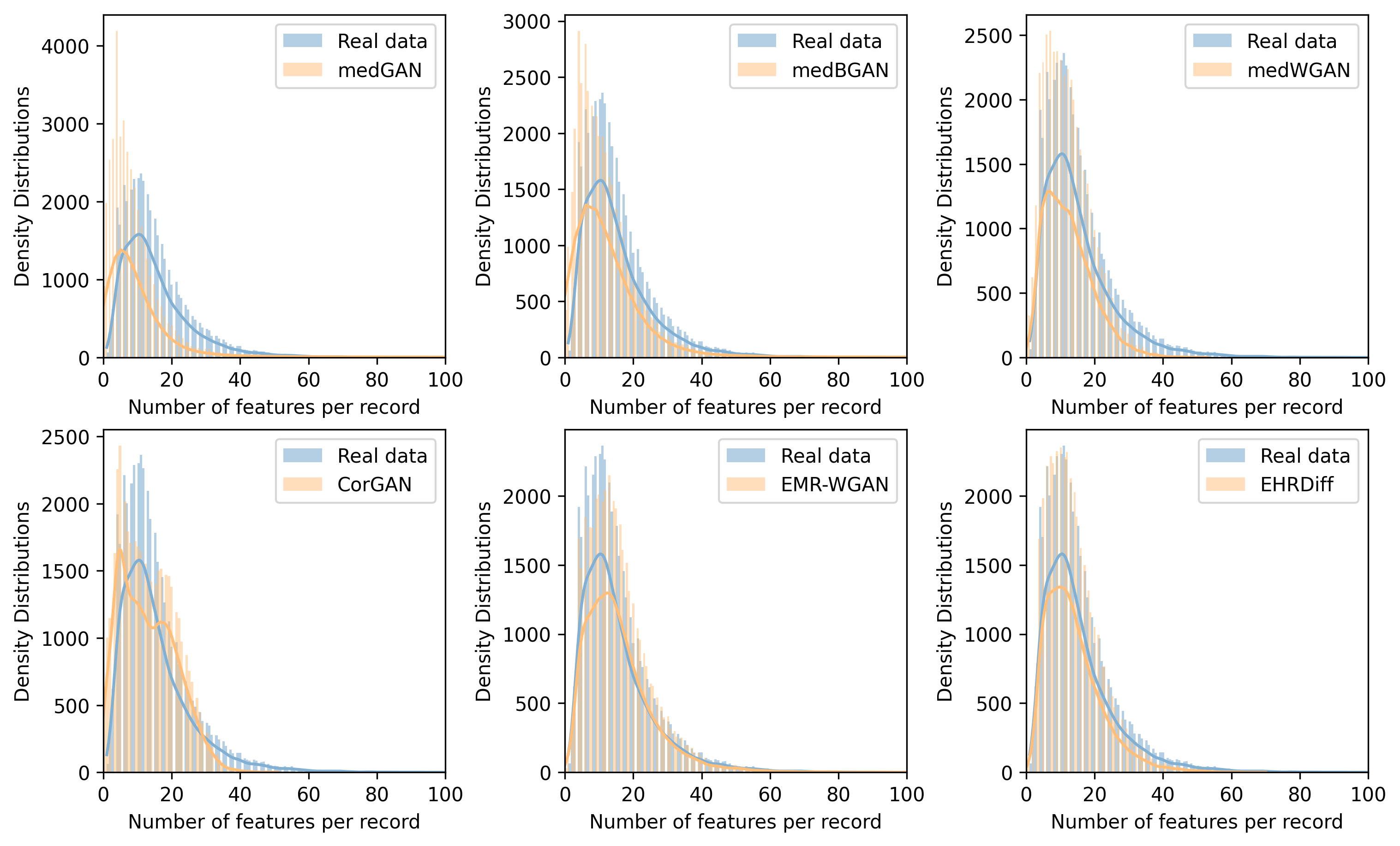}
    \end{center}
    \caption{The histograms plot the empirical distributions of the unique code counts on the sample level. The solid lines are the kernel density estimations of the distribution.}
    \label{fig: histo}
\end{figure*}

The quality of synthetic EHR data can be evaluated from a multifaceted perspective \citep{Yan2022AMB}. We use additional metrics to further evaluate the synthetic EHR data on MIMIC-III.

\subsection{Latent cluster analysis} 
The metric evaluates the distributional difference between the synthetic and real EHR data in the latent space. The metric first uses principle component analysis to reduce the sample dimension for both data and then cluster the samples in the latent space. Ideally, if synthetic and real EHR data are identically distributed, the synthetic and real EHR samples should respectively comprise half of the samples in one cluster. Therefore, the metric is calculated as:
\begin{align}
    \log\left(\frac{1}{K}\sum_{i=1}^K[\frac{n_i^{\text{real}}}{n_i}-0.5]^2\right),
\end{align}
where $K$ is the number of resulted clusters, $n_i$ and $n_i^{\text{real}}$ are the sample number and the real sample number in $i$th cluster, respectively. The lower the value, the less synthetic data distribution deviates from the real data distribution. In our experiments, $K$ is decided by the elbow method \citep{yuan2019research} for each synthetic data and in this work is 4 or 5 according to different methods.

\subsection{Medical concept abundance} 
The metric assesses the synthetic EHR data distribution on the record level. The metric calculates the empirical distribution of the unique positive (occurred) code number within each sample. The empirical distributions are calculated by histograms. The discrepancy between synthetic and real EHR data is calculated as follows:
\begin{align}
    \sum_{i=1}^M\frac{1}{2N}|h_r(i)-h_s(i)|,
\end{align}
where $M$ is the number of bins in histograms, $N$ denotes the number of samples for real (or synthetic) data, and $h_r(i)$ and $h_s(i)$ respectively represent the $i$th bin in the histograms of real and synthetic EHR data. In this work, M is set to $20$.

\subsection{Results}

From Table \ref{tabapp:utility}, it is shown that \modelname performs better than most baselines and only marginally falls behind EMR-WGAN by 0.877 on the latent distance metric. In terms of MCAD, \modelname consistently outperforms all baselines, and as depicted in Figure \ref{fig: histo}, we can see that the histogram of unique code count distribution of synthetic EHR data by \modelname achieves the best fit to that of real EHR data. From a sample-level perspective, latent distance results illustrate that synthetic EHR data by \modelname is closely distributed to real EHR data. The MCAD results show that synthetic data by \modelname resembles the real EHR most in terms of unique positive code counts. This result is in line with the findings of the non-zero code column metric.

\section{Data Materials}
\label{app:data}

\subsection{CinC2012 Data} 
CinC2012 Data \citep{silva2012predicting} is a dataset proposed to predict the mortality of ICU patients in the CinC2012. It contains general descriptors such as age, gender, and ICU type and time series records like heart rate, respiration rate, and serum glucose. In our experiments, we use the preprocessed version of this dataset from \citep{johnson2012patient}, which is derived by applying simple extraction on the time-series features and excluding abnormal outliers in the physiological measurements. We then added the label of in-hospital mortality to the records, making 115 features in total. There are 4000 records for model training and another 4000 records for model testing, as split by the CinC authority. We use this dataset to evaluate the models' performance on mixed-type EHR data.

\subsection{PTB-ECG Data}
PTB-ECG Data \citep{bousseljot1995nutzung} is a collection of ECG signals for heart disease diagnosis. We utilize a preprocessed version from \citep{kachuee2018ecg} to carry on our experiments, where the signals are segmented and preprocessed from the original PTB Diagnostic ECG Database. The dataset contains 4046 normal patients and 10506 records with heartbeat classified as abnormal. Specifically, all the signals are cropped, downsampled, or padded to make each sample into a fixed dimension of 188. We use this dataset to explore models' ability to generate continuous medical time series data.

\end{document}